\documentclass[final,3p,times,twocolumn]{elsarticle}
\usepackage{amssymb}
\usepackage{mathtools, graphicx}
\usepackage{multirow}
\usepackage{slashbox}
\usepackage{amssymb}
\journal{Pattern Recognition}

\begin{document}

\begin{frontmatter}


\title{Modelling Temporal Information Using Discrete Fourier Transform \\for Video Classification}

%\author{Haimin Zhang and Min Xu}
\author[uts]{Haimin Zhang}
%\author[uts]{Min Xu\corref{cor1}} %
\ead{Min.Xu@uts.edu.au}
%\author[cas]{Changsheng Xu}
%\author[uci]{Ramesh Jain}

\cortext[cor1]{Corresponding author}

\address[uts]{Faculty of Engineering and IT, University of Technology Sydney}%\fnref{label3}}, 15 Broadway, NSW 2007, Australia
%\address[cas]{National Lab of Pattern Recognition, Institute of Automation, Chinese Academy of Sciences}%, 95 Zhongguancun East Road, Beijing 100190, P.R.China
%\address[uci]{Bren School of Information and Computer Sciences, University of California, Irvine}%, Irvine, CA 92697, United States

\begin{abstract}
%% Text of abstract\textbf{[NO MORE THAN 150 WORDS CURRENTLY 194]} With the widespread of user-generated Internet videos,
Recently, video classification attracts intensive research efforts. However, most existing works are based on frame-level visual features, which might fail to model the temporal information, e.g. characteristics accumulated along time. In order to capture video temporal information, we propose to analyse features in frequency domain transformed by discrete Fourier transform (DFT features). Frame-level features are firstly extract by a pre-trained deep convolutional neural network (CNN). Then, time domain features are transformed and interpolated into DFT features. CNN and DFT features are further encoded by using different pooling methods and fused for video classification. In this way, static image features extracted from a pre-trained deep CNN and temporal information represented by DFT features are jointly considered for video classification. We test our method for video emotion classification and action recognition. Experimental results demonstrate that combining DFT features can effectively capture temporal information and therefore improve the performance of both video emotion classification and action recognition.
% performance.
Our approach has achieved a state-of-the-art performance on the largest video emotion dataset (VideoEmotion-8 dataset) and competitive results on UCF-101.
%, improving accuracy from 51.1\% to 62.6\%.
\end{abstract}


\end{frontmatter}

\section{Introduction}
\label{sec:intro}
\begin{figure*}[!htd]
\begin{center}
\includegraphics[width=\textwidth]{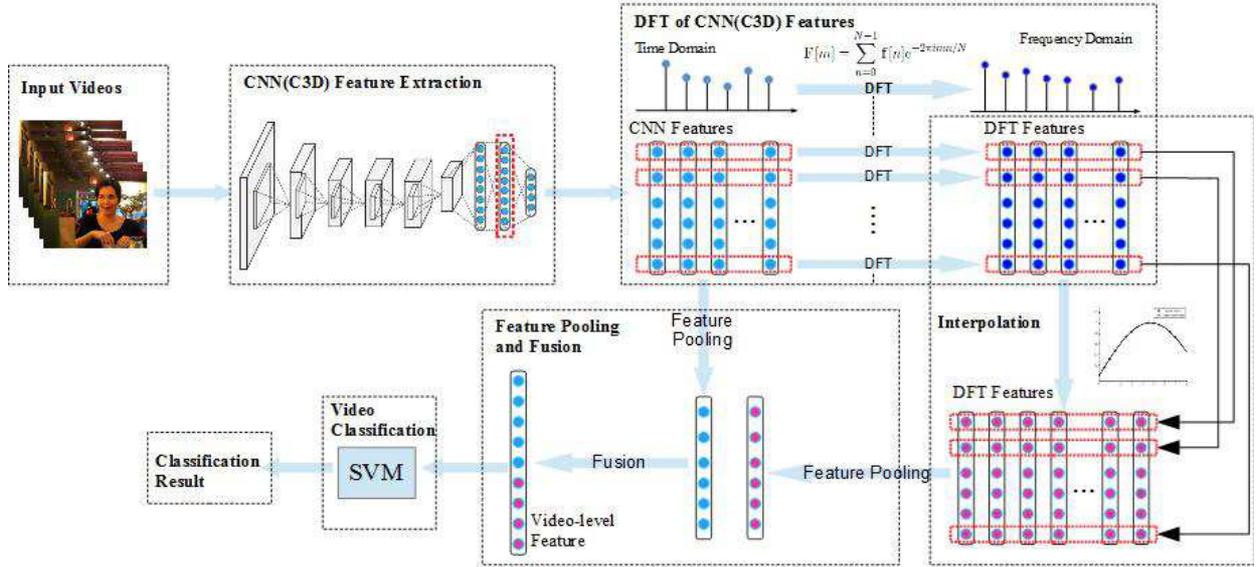}
\caption{An overview of our method for video classification. Given an input video, CNN(C3D) features are firstly extracted. Then these features are transformed to frequency domain using DFT. Feature pooling methods are applied for CNN(C3D) feature and DFT features separately. The concatenation of the aggregated CNN features and DFT features lead to the video-level representation.  Finally, an SVM is trained to differentiate different classes of video categorises.}
\end{center}
\end{figure*}
As technology advances, millions of video clips are uploaded to the Internet every day. From video sharing websites (e.g. YouTube and Flicker), people can easily access hundreds of video clips. It is an essential and urgent demand to develop intelligent algorithms for classifying these videos, which form the basis of various potential applications, such as video retrieval, recommendation and annotation.

Video classification is to automatically assign a label to a video clip. Recent research classify video sequences into either emotional classes \cite{Jiang2014,pang2015mutlimodal, pang2015deep} or action categories \cite{simonyan2014two, srivastava2015unsupervised,tran2014learning}.  Compared to static image classification, video classification is a complex task. This is mainly because of three reasons: 1) Video classification demands high computation cost since a short video can contain hundreds of or even thousands of frames. 2) A video sequence contains multiple frames which are grouped together to reflect a dominant theme. Within a video sequence, some frames might not convey (or even convey an opposite meaning of) the dominant theme. 3) Spatial and temporal information should be jointly considered for video classification. Static image features have been extensively studied by researchers over the decades. Static image features can be utilised to represent video at a frame-level. In addition to static image features, temporal information is an import clue for video classification. Since the length of video clips are different, it is a challenging task to generate uniform video representations which can jointly model spatial and temporal information.

Most conventional approaches of video classification involve mainly three stages: First, local features or frame-level features are extracted. Then these features are quantised to a fixed length representation using a visual dictionary usually learned by K-means algorithm. Later, video-level representations are obtained by pooling method, such as max-pooling and average-pooling. Lastly, a classifier is trained on the video-level representations to differentiate different classes of video categorises.
%Considering one of the difficulties as mentioned above, Unlike static image emotion analysis, video has temporal information.

However, during the process of feature quantisation and pooling, the temporal information (\emph{i.e.}, information accumulated along time) of videos has not been well treated. Without modelling temporal information, the performance of classifiers are restricted.
Taking activity recognition as an example, without considering temporal information, a classifier cannot differentiate activities of standing up v.s. sitting down and opening doors v.s. closing doors.

Most recently, realising the importance of temporal information, researchers have started exploring how to take temporal information into account for video classification. One dominated approach in recent years is Dense Trajectories \cite{wang2011action}, which tracks densely sampled image patches over time and calculates conventional local features, e.g. HOG \cite{dalal2005histograms}, HOF \cite{laptev2008learning} and MBH \cite{dalal2006human}, based on the calculated trajectories. However, the application of Dense Trajectories is restricted to relatively small scale dataset since the method needs intensive computation.

In \cite{wu2015modeling}, Long Short Term Memory (LSTM) \cite{hochreiter1997long}, which can preserve information for a long time, was adopted to model temporal information for video classification. However, training an LSTM is a time-consuming task.
Although these initiatives can somehow capture temporal information, they might failed to consider the accumulated information along time.

Inspired by Discrete Fourier Transform (DFT), which can transform a discrete signal form time domain to frequency domain, this paper proposes to analysis features in frequency domain to model temporal information for videos. To some extent, signal characteristics along time can be accumulated and represented through sampling in frequency domain \cite{rader1968discrete}. In this work, we integrate CNN features (features extracted from a pre-trained convolutional neural network) and DFT features for video representation by leveraging feature encoding method.

%??????????????????????????????????????????
The proposed method consists of five steps as shown in Figure 1. The first step is to extract CNN or C3D (convolutional 3D) features from videos. Secondly, considering each feature dimension as a discrete signal over time, we apply DFT to transform the signal to frequency domain.
Thirdly, an interpolation method is adopted to generate a fixed length representation for every dimension of DFT features.
Fourthly, feature encoding methods (i.e. average pooling, LLC, FV and VLAD) are applied to aggregate CNN features and DFT features. The combination of the aggregated CNN features and DFT features can be regarded as a video-level representation. Finally, with the features generated in the fourth step, an SVM is trained for video classification.
Different from existing methods for temporal information analysis, our method has the following advantages:
\begin{itemize}
\item In order to capture video temporal information, we propose to analyse features in frequency domain transformed by discrete Fourier transform.
%Compared to LSTM, our method avoids a complex model learning process.
\item Video clips having different length places a challenge for obtaining a uniformed feature representation. Using DFT is resilient to signal length variation. Moreover, compared to DFT, applying fast Fourier transform (FFT) can significantly reduce the computational complexity.
 %We test our method for video emotion classification and action recognition. Experimental results demonstrate that combining DFT features can effectively capture temporal information and therefore improve the performance of both video emotion classification and action recognition. Our approach has achieved a state-of-the-art performance on the largest video emotion datase.
\item Experiments on two tasks, i.e. video emotion classification and action recognitions, demonstrate that combining DFT features can effectively capture temporal information and therefore improve the performance of video classification.
%Our method effectively captures temporal information of videos by analysing features in frequency domain.% with relatively low feature dimensions.
%
\end{itemize}
%??????????????????????????????????????????

The remainder of this paper is organised as follows. In section 2, we review the related work on video representations using temporal information. In section 3, DFT-based temporal information modelling is introduced in details. Experimental results are presented and discussed in section 4. Finally, we conclude this paper in section 5.

\section{Related Work}
\subsection{Low-level Representations}
Although video classification has been researched for many years, it is still a challenging task that attracts much research interests over the decades. Early works on video classification focus on employing efficient image representations. Local image features (such as SIFT \cite{lowe2004distinctive}, HOG (histograms of oriented gradients) \cite{dalal2005histograms} and HOF (histograms of optical flow) \cite{laptev2008learning}) are extracted from video frames. These features are further encoded to generate a uniform video-level representation. As one of the feature encoding method, bag of visual words (BoVW) \cite{yang2007evaluating} is the most commonly used model for generating a uniform representation.
%???????????????????????????????????????
%In [?], sparse coding was proposed to ...... In [?] (LLC) Other feature encoding methods sparse coding, LLC encoding, fisher vector (FV) and (VLAD).
Sparse coding, which uses sparse constraints, is an extension of the BoVW model. Sparse coding achieves less quantisation error than BoVW. In \cite{yang2009linear}, sparse coding was proposed for image classification. In addition to sparse constraints, LLC utilises locality constraints and can further reduce quantisation error. In \cite{wang2010locality}, LLC was proposed for image classification. Fisher vector, which was derived from fisher kernel \cite{perronnin2007fisher} , was firstly introduced for large-scale image classification in \cite{perronnin2010improving}. However, the dimension of features generated by FV is much higher than BoVW and  LLC. % VLAD can be viewed as a simplification of F
Vector of locally aggregated descriptors (VLAD) is anther popular feature encoding approach, which was firstly proposed by Jegou in \cite{jegou2010aggregating} for image representation.
%???????????????????????????????????

\subsection {High-level Representations}
%?????????????????????
Due to the limited discriminative capacity of low-level representations to video semantics, high-level representations were introduced.
In high-level feature representations, an image is represented as a response map of a large number of pre-trained detectors. %, were proposed for visual task analysis.
In \cite{li2010object}, Object Bank was proposed for scene classification and semantic feature sparsification. SentiBank, which consists of 1,200 concepts and associated classifiers, was constructed for sentiment prediction in images in \cite{borth2013sentibank}. Action Bank, which is comprised of many individual action detectors sampled in semantic space and viewpoint space, was proposed for action recognition in \cite{sadanand2012action}.
Jiang \emph{et. al.} \cite{Jiang2014} applied ObjectBank \cite{li2014object} and Sentibank \cite{borth2013sentibank} for video emotion categorisation.
%????????????????????????

\subsection{Deep-learned Features}
In contrast to hand-crafted features, the last few years have witnessed the success of deep features. Deep features extracted from the activation of a convolutional neural work (CNN) pre-trained on a large image dataset (e.g. ImageNet \cite{deng2009imagenet}) have proved to be more discriminative than hand-crafted features \cite{jia2014caffe}. CNN features have achieved state-of-the-art results on many benchmarks \cite{razavian2014cnn} and are widely used in image classification, object detection and attribute detection \cite{zhou2014learning, razavian2014cnn}.
Recently, researchers started applying CNN features to video classification \cite{simonyan2014two, wu2015modeling, xu2014discriminative}. %Simonyan \emph{et. al.} \cite{simonyan2014two} proposed a two-stream architecture for video classification using CNN features and motion features.
Xu \emph{et. al.} \cite{xu2014discriminative} proposed a video representation method through leveraging frame-level features extracted by CNN with FV and VLAD as feature encoding method. In \cite{simonyan2014two}, two stream convolutional networks was proposed on the top of video frames and stacked optical flows to capture spatial and motion information.

\subsection {Temporal Information Modelling}
%?????????????????
Researchers started to model temporal information using motion features. Dense Trajectories \cite{wang2011action}, which was inspired by dense sampling method in image classification, was proposed for action recognition.
%Dense Trajectories (DT)....... [?]
%????????????
As a derived version of DT, improved Dense Trajectories (IDT) \cite{wang2013action} were proposed to improve the performance of DT by taking camera motion into consideration and tested on a number of challenging dataset (e.g. HMDB51 \cite{kuehne2011hmdb}, Sports-1M \cite{karpathy2014large}). However, motion features can only capture temporal information over a couple of consecutive frames.
As a special architecture of recurrent neural network (RNN), long short term memory (LSTM) was specifically designed with memory cells to store, modify and access its internal states, and can persist long time information.
LSTM was successfully used to capture temporal information for sequence learning tasks, such as  speech recognition \cite{al2009network} and machine translation \cite{sutskever2014sequence}. In\cite{srivastava2015unsupervised}, an unsupervised learning approach was proposed for video representation using the LSTM encoder-decoder architecture.

Conventional CNN are only limited to handle frame-level inputs. As an extension of CNN,  3D convolutional neural network was proposed for videos classification in \cite{ji20133d}, which extract features from both spatial and temporal dimensions by performing a 3D convolution and pooling. In \cite{tran2014learning}, a 3D convolutional neural network was trained to extract spatial-temporal features. However, 3D CNN architecture can only take video clips with a short length (usually, 16 frames) as inputs. This stops it from capturing long term accumulated temporal information.

\section{The proposed DFT-based Temporal Information Modelling}
The proposed algorithm involves five steps and is introduced in details step by step in this section.
\subsection{Feature Extraction}
\textbf{CNN Features}
As shown in \cite{razavian2014cnn}, deep features extracted from a convolutional neural network which is pre-trained on a large image dataset can be used as a powerful feature representation for many visual analysis tasks. In this paper, we leverage a deep convolutional neural network \cite{krizhevsky2012imagenet} pre-trained on ImageNet \cite{deng2009imagenet}, which contains 1.2 million images categorised into 1000 classes, to extract frame-level descriptors for all video clips.
The network consists of five convolution layers and three fully connected layers with a final 1000-way softmax. All input images are resized to 256$\times$256 without considering its original aspect ratio before feeding into the network. Considering features extracted from fully connected layers can capture semantic information from the input image, activation from the fully connected layer are extracted as the frame descriptor. Following \cite{xu2014discriminative}, fc$_6$ and fc$_7$ refer to the activation of the first and second fully-connected layers. Then \ensuremath{\ell_2} normalisation is adopted to all frame-level descriptors.

\textbf{C3D Features}
C3D (Convolution 3D) features refers to features extracted from a pre-trained 3D convolutional networks \cite{tran2014learning}. Unlike convolutional networks, 3D convolutional networks takes a short video clip (usually 16 frames) as input and leverages on 3D convolution and pooling.%, and thus can learn motion information.

Let $\mathbf{f}$ denote the set of frame-level descriptors of a video clip which has $N$ frames, then $\mathbf{f}$ can be described as
\[
    \mathbf{f}=(\mathbf{f}_1,...,\mathbf{f}_N) =
    \left[
      \begin{array}{cccccc}
        \mathbf{f}_1[1]   & \hdots &\mathbf{f}_i[1] & \hdots & \mathbf{f}_N[1] \\
        \vdots  & \vdots  & \vdots  &  \vdots  & \vdots\\
        \mathbf{f}_1[k]   & \hdots &\mathbf{f}_i[k] & \hdots & \mathbf{f}_N[k] \\
        \vdots  & \vdots  & \vdots  &  \vdots  & \vdots\\
        \mathbf{f}_1[D]   & \hdots &\mathbf{f}_i[D] & \hdots & \mathbf{f}_N[D] \\
      \end{array}
    \right]
    \]
Where $\mathbf{f}_i=(\mathbf{f}_i[1],...,\mathbf{f}_i[D])^{T}$ represents the descriptor of the $i$-th frame with dimension $D$. In this work, $D$ equals 4096, which is the dimension of fc$_6$. The value of $N$ can be different for different video clips.

\subsection{Discrete Fourier Transform of CNN Features}
The aim of discrete Fourier transform (DFT), which is widely used in the field of signal processing, is to transform a discrete signal from time domain to frequency domain. At this step, we present how DFT is applied to CNN features.

As described in 3.1, $\mathbf{f}=(\mathbf{f}_1,...,\mathbf{f}_N)$ represents the set of frame-level CNN features extracted from a video clip with $N$ frames.
Let the $k$-th dimension of $\mathbf{f}$ be denoted $\mathbf{f}[k]=(\mathbf{f}_1[k],\mathbf{f}_2[k],...\mathbf{f}_N[k])$. $\mathbf{f}[k]$ can be considered as a discrete signal which has $N$ sample points with equal sampling time intervals $\Delta t$. We transform $\mathbf{f}[k]$ to frequency domain using the following equation
\begin{equation} \label{eq:DFT}
    \begin{aligned}
     \mathbf{F}_s[k] = \sum_{n=1}^{N}\mathbf{f}_n[k]e^{-2i \pi (n-1)(s-1)/N}, \; s=1,2,...,N.
    \end{aligned}
\end{equation}

Let the result be denoted $\mathbf{F}[k]=(\mathbf{F}_1[k],...,\mathbf{F}_N[k])$, the number of points obtained in frequency domain is as same as that in time domain.
The computed value $\mathbf{F}_s[k] \in \mathbf{F}[k]$ is a complex number and its absolute value represents the amplitude of the $s$-th frequency. In our work, the absolute value of $\mathbf{F}_s[k]$ is used instead of its original complex value. After transforming all $\mathbf{f}[k]$ to frequency domain, where $k=1,...,D$, we get the following feature set
\[
    \mathbf{F}=(\mathbf{F}_1,...,\mathbf{F}_N) =
    \left[
      \begin{array}{cccccc}
        \mathbf{F}_1[1]   & ... &\mathbf{F}_i[1] & ... & \mathbf{F}_N[1] \\
        \vdots  & \vdots  & \vdots  &  \vdots &  \vdots\\
        \mathbf{F}_1[k]   & ... &\mathbf{F}_i[k] & ... & \mathbf{F}_N[k] \\
        \vdots  & \vdots  & \vdots  &  \vdots &  \vdots\\
        \mathbf{F}_1[D]   & ... &\mathbf{F}_i[D] & ... & \mathbf{F}_N[D] \\
      \end{array}
    \right]
    \]
 Where $\mathbf{F}_i=(\mathbf{F}_i[1],...,\mathbf{F}_i[D])^{T}$, termed as a DFT feature in this paper.

\subsection{Interpolation}
%The difference is that for long videos, we sample more points in the frequent domain, this can be shown in figure 2.
%More points can be obtained using interpolation method. In the experiment section, we will compare results using interpolation and without using interpolation.
%Let the Fourier Transform of the original signal $f(t)$ be denoted $\mathcal{F}f(s)$.
%$F[0],F[1],...,F[N-1]$ are points evaluated at $0,1/L,...,(N-1)/L$ and provide an approximation of $\mathcal{F}f(0),\mathcal{F}f(1/L),...,\mathcal{F}f((N-1)/L)$. The number of sample points $N$ can be represented as:
%\begin{equation}
%    \begin{aligned}
%     N = \frac{B}{1/L}=BL
%    \end{aligned}
%\end{equation}
%Due to the large length variances of the user-generated videos, different video generally has different time period, so the number of frames extracted from different videos is usually different from each other.
As mentioned in section 3.1, the number of sample points $N$ is different duo to the various video length.
For two video clips $u$ and $v$ with $N$ and $M$ frames respectively, let $\mathbf{f}^u[k]=(\mathbf{f}^u_1[k],\mathbf{f}^u_2[k],...,\mathbf{f}^u_N[k])$ and $\mathbf{f}^v[k]=(\mathbf{f}^v_1[k],\mathbf{f}^v_2[k],...,\mathbf{f}^v_M[k])$ indicate the $k$-th dimension of CNN features.
We use $\Delta t$ to indicate the sampling time interval which is uniform for all video clips, \emph{i.e.} the sampling rate is $S=1/\Delta t$.

After transforming $\mathbf{f}^u[k]$ and $\mathbf{f}^v[k]$ to frequency domain, we obtian $\mathbf{F}^u[k]=(\mathbf{F}^u_1[k],\mathbf{F}^u_2[k],...,\mathbf{F}^u_M[k])$ and $\mathbf{F}^v[k]=(\mathbf{F}^v_1[k],\mathbf{F}^v_2[k],...,\mathbf{F}^v_M[k])$ respectively.

$\mathbf{F}^v[k]$ and $\mathbf{F}^v[k]$ have the same frequency range from 0 to $S$ with sampling interval $S/N$ and $S/M$ respectively.

From equation \ref{eq:DFT}, we know that the number of points obtained in frequency domain is as same as that in time domain. Therefore signals with more sample points in time domain are more compactly spaced in frequency domain than signals with less sample points.
\begin{figure}[!htbp]
\begin{center}
\includegraphics[width=\textwidth/2-1cm]{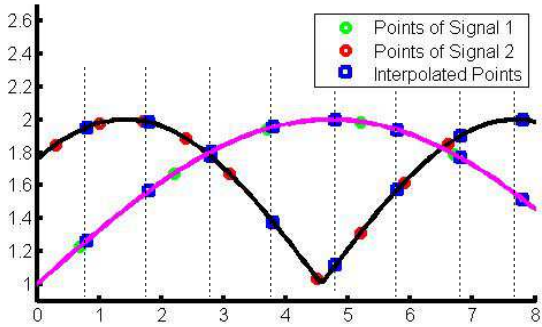}
\caption{Illustration of applying interpolation to signals with different points in frequency domain.}
\end{center}
\end{figure}

Like image resizing, we use cubic interpolation method \cite{keys1981cubic} to generate a fixed length ($L$) representation $\mathbf{F}^u[k]=(\mathbf{F}^u_1[k],\mathbf{F}^u_2[k],...,\mathbf{F}^u_L[k])$ and $\mathbf{F}^v[k]=(\mathbf{F}^v_1[k],\mathbf{F}^v_2[k],...,\mathbf{F}^v_L[k])$. By this way, different video clips have the same frequency sample interval from the frequency range from 0 to $S$, as shown in Figure 2.

\subsection{Feature Pooling and Fusion}
So far, the CNN features are extracted at frame-level. In order to generate a uniform video-level representation, we need to aggregate the obtained CNN features and DFT features separately.
%LLC, FV and VLAD have been proved to outperform BoVW model. Therefore, we apply the above-mentioned three encoding methods to
We apply four most commonly used pooling methods, i.e. average pooling, locality-constrained linear coding (LLC), Fisher vector (FV) and vector of locally aggregated descriptors (VLAD) to aggregate both CNN features and DFT features in our experiments, which are briefly reviewed in this section.
\subsubsection{Average Pooling}
Average pooling is simply to calculate the mean value of the feature vector. Suppose that $\mathbf{f}=(\mathbf{f}_1,...,\mathbf{f}_N)$ represent the set of frame-level features extracted from a video clip, the video-level features generated using average pooling can be represented as
\begin{equation} \label{eq:VLAD}
\begin{aligned}
    %\tilde{\textbf{x}}=\frac{1}{N}\sum_{i=1}^{N}\textbf{x}_i
    {\textbf{f}_{\rm{avg}}}=\frac{1}{N}\sum_{i=1}^{N}\textbf{f}_i
\end{aligned}
\end{equation}
%Video-level features are then re-normalised.

The dimension of the video-level features generated by average pooling is same as the frame-level features. The calculation of average pooling is easy. The disadvantage is that the temporal information between frames is totally lost.

\subsubsection{LLC Encoding}
LLC, which utilises locality constraints, selects k-nearest codewords from a dictionary learned by K-means algorithm, and generates a sparse representation for the input vector. Given an input vector $\textbf{x}$, which can be a frame-level CNN feature or a DFT feature in our case, LLC code can be obtained by solving the following fitting problem:
\begin{equation} \label{eq:VLAD}
\begin{aligned}
& \min_{\textbf{c}} \sum_{i=1}^{N}\| \textbf{x}-\textbf{Bc} \|^2+\lambda \|\textbf{d}_i \odot \textbf{c}\|\\
& \;\;\;\;\;s.t. \; \textbf{1}^T\textbf{c} = 1
\end{aligned}
\end{equation}
where $\odot$ denotes element-wise multiplication, and $\textbf{d}_i$ is the locality adaptor. The solution of LLC can be derived analytically \cite{wang2010locality} by:
\begin{equation}
%\tilde{\textbf{c}}=(\textbf{C}_i+\lambda \rm{diag(d)})\;\backslash \;\textbf{1}\\
\textbf{c}=\tilde{\textbf{c}}/\textbf{1}^T
\end{equation}
where $\tilde{\textbf{c}}=(\textbf{C}_i+\lambda \rm{diag(\textbf{d})})\backslash\textbf{1}$ and $\textbf{C}_i=(\textbf{B}-\textbf{1x}_i^T)(\textbf{B}-\textbf{1x}_i^T)^T$.
After that, max-pooling strategy is applied to aggregate LLC-based features.

\subsubsection{Fisher Vector Encoding}
Fisher vector representation does not require as many visual words as LLC. In Fisher vector \cite{sanchez2013image,xu2014discriminative} encoding, the vocabularies of visual words are represented by the means of a Gaussian mixture model (GMM), which is learned in an unsupervised manner. Let a GMM model with $K$ components be denoted as $\Theta=\{(\mu_k,\Sigma_k,\pi_k),k=1,2,..,K\}$, where $\mu_k$, $\Sigma_k$ and $\pi_k$ represent the mean, variance and prior parameter of the $k$-th component, respectively. Let %$X=(\textbf{x}_1,...,\textbf{x}_N)$ denote a set of frame-level CNN descriptors extracted from a
$\mathbf{f}=(\mathbf{f}_1,...,\mathbf{f}_N)$ denote the set of frame-level CNN descriptors extracted from a video clip with $N$ frames, then the mean and covariance deviation vectors for the $k$-th component are computed as:
\begin{equation} \label{eq:FV}
    \begin{aligned}
     \textbf{u}_k = \frac{1}{N\sqrt{\pi_k}} \sum_{i=1}^{N} q_{ki}(\frac{\textbf{f}_i-\mu_k}{\sigma_k})\\
     \textbf{v}_k = \frac{1}{N\sqrt{\pi_k}} \sum_{i=1}^{N} q_{ki}(\frac{\textbf{f}_i-\mu_k}{\sigma_k})
    \end{aligned}
\end{equation}
where $q_{ki}$ represents the posterior probability. The concatenation of $\textbf{u}_k$ and $\textbf{v}_k$ of all the $K$ components lead to the final Fisher vector representation.

The dimension of video-level features generated by Fisher vector is $2DK$, where $D$ indicates the dimension of frame-level CNN descriptor. %The dimension of vectors generated by FV is much higher than that of  generated by LLC.
\subsubsection{VLAD Encoding}
Vector of locally aggregated descriptors (VLAD) \cite{jegou2010aggregating} can be viewed as a simplification of the Fisher vector representation. Same as LLC encoding, a visual dictionary $\mathcal{C}=\{c_1,...,c_K\}$ of $K$ visual words is learned by K-means method. Let $\mathbf{f}=(\mathbf{f}_1,...,\mathbf{f}_N)$ denote the set of frame-level CNN descriptors extracted from a video clip with $N$ frames. Each vector $\mathbf{f}_i$ is associated with its nearest visual word $c_i=\rm{NN}(\mathbf{f}_i)$. Different vector regarding center $\textbf{c}_k$ can be obtained by:
\begin{equation} \label{eq:VLAD}
    \begin{aligned}
     \textbf{u}_k = \sum_{i:\rm{NN}(\textbf{x}_i)=\textbf{c}_k}(\textbf{x}_i-\textbf{c}_k)
    \end{aligned}
\end{equation}
The dimension of video-level features generated by VLAD is $KD$, where $D$ represents the frame-level feature dimension. Compared with FV, the cost for calculating VLAD can be significantly reduced.
%$\sum_{i=1}^n a_i=0$
\subsubsection{Feature Fusion}
After obtaining aggregated CNN features and DFT features. We then adopt late feature fusion. The linear combination of the aggregated CNN features and DFT features lead to the final video-level representation, which is denoted as $\tilde{\textbf{x}}$.
%In one of the later sections, we will explore different weighted combination of CNN and DFT features and discuss the comparison results.

\subsection{Video Classification}

After obtaining all video-level features, an SVM is trained by optimising the following equation \cite{fan2008liblinear} for video classification.
\begin{equation} \label{eq:SVM}
\mathrm{arg}\,\mathrm{\min_{\textbf{w},b}}\; \frac{1}{2} \|\textbf{w}\|^2+C\sum_{i=1}^{T}max(1-y_i(\textbf{w}^T {\textbf{x}}_i) + b),0)
%\mathrm{arg}\,\mathrm{\min_{\textbf{w},b}}\; \frac{1}{2} \|\textbf{w}\|^2+C\sum_{i=1}^{T}max(1-y_i(\textbf{w}^T \tilde{\textbf{x}}_i) + b),0)
\end{equation}
Where ${\textbf{x}}_i$ and $y_i$ represent video-level feature and its corresponding label, $C$ and $T$ indicate penalty parameter and the number of training features respectively.

\section{Experiments}
Recently, video emotion classification and action recognition have attracted intensive research efforts. In this section, in order to evaluate the effectiveness of the proposed DFT-based temporal information model, two sets of experiments were conducted: video emotion classification and action recognition.
\subsection{Video Emotion Classification}
%To verify the effectiveness of the proposed algorithm, experiments are conducted on two public datesets: VideoEmotion-8 \cite{Jiang2014} and UCF-101 \cite{soomro2012ucf101}.
In our experiments, we intended to evaluate: (1) the performance of CNN features with different pooling methods; (2) the performance of DFT features with different pooling methods; and (3) the overall performance of combined CNN features (with different pooling methods) and DFT features (with different pooling methods). Moreover, to prove the efficiency of the proposed method, we compared our results with the most recent three works \cite{Jiang2014}, \cite{pang2015mutlimodal} and \cite{pang2015deep}.
\subsubsection{Video Emotion Dataset}
\textbf{VideoEmotion-8 dataset} It contains 1,101 user-generated videos labelled with 8 basic human emotion categories. There are at least 100 videos in each category.  The average duration of the 1,101 videos is 107 seconds. Currently, this is the largest dataset available for recognising emotions in user-generated videos. These videos were collected from popular video sharing websites, \emph{i.e.} Youtube and Flicker.
Similar as \cite{Jiang2014}, we randomly selected $2/3$ data from each category for training and the rest for testing. Experiments were conducted ten times. The average accuracy of the ten times was calculated to evaluate the classification performance.
For computation efficiency, we sampled a frame every 15 frames.

\subsubsection{Implementation Details}
%the activations of fc$_6$ are extracted as frame-level features, followed by \ensuremath{\ell_2} normalisation.
The activation of fc$_7$ were extracted as frame-level features
using the Caffe toolkit \cite{jia2014caffe}, and were further \ensuremath{\ell_2} normalised. We implemented LLC according to \cite{wang2010locality}. FV and VLAD representation were generated by utilising vlfeat \cite{vedaldi2010vlfeat}.
%Learning spatiotemporal features with 3d convolutional networks

Fast Fourier transform (FFT) \cite{frigo1998fftw} was adopted to compute DFT. The dimension of CNN features were reduced from 4,096 to 1,024 using principal component analysis (PCA). At the interpolation step, $L$ (mentioned in section 3.3) was set to 500 experimentally. For CNN and DFT features, different feature pooling methods were applied and compared. We trained a vocabulary with 1,024 codewords for LLC, 16 codewords for FV, and 16 codewords for VLAD. In our experiments, the aggregated CNN features were normalised to $3/5$ and the aggregated DFT features were normalised to $2/5$.

We applied the LibLinear toolbox \cite{fan2008liblinear} for SVM classification. The penalty parameter $C$ was set to 100 experimentally.

\begin{table*}[!htbp]
\centering  % 表居中
\begin{tabular}{|c|c|c|c|c|c|c|c|c|c|}  % {lccc} 表示各列元素对齐方式，left-l,right-r,center-c
\hline
Method &Anger &Anticipation &Disgust &Fear &Joy &Sadness &Surprise &Trust &Overall\\
\hline
\hline
$\rm{CNN}_{Avg}$   &56.8	&42.5	&58.4	&67.4	&68.5	&64.5	&78.3	&40.3	&-1\\
$\rm{CNN}_{LLC}$   &60.2	&46.9	&55.6	&68.3	&62.6	&57.1	&72.4	&48.7	&-1\\
$\rm{CNN}_{FV}$    &48.5	&71.6	&68.4	&76.9	&64.3	&67.0	&75.5	&51.8	&\textbf{-1}\\
$\rm{CNN}_{VLAD}$  &44.1	&49.4	&70.1	&65.1	&54.8	&64.8	&55.3	&56.4	&-1\\
\hline
\hline
$\rm{DFT}_{Avg}$   &42.1	&15.3	&32.4	&54.4	&48.8	&47.3	&69.5	&20.3	&-1\\
$\rm{DFT}_{LLC}$   &33.7	&15.8	&41.9	&38.3	&45.8	&50.0	&66.2	&23.0	&-1\\
$\rm{DFT}_{FV}$    &68.2	&37.2	&42.9	&53.0	&33.3	&68.2	&65.6	&33.3	&\textbf{-1}\\
$\rm{DFT}_{VLAD}$  &46.5	&12.5	&35.8	&55.2	&42.5	&57.6	&72.3	&30.0	&-1\\
\hline
\end{tabular}
\caption{Prediction accuracy (\%) of each emotion category using CNN and DFT features on VideoEmotion-8 dataset.}
\end{table*}

%%%%%%%%%%%%%%%%%%%%%%%%%%%%%%%%%%%%%%%%%%%%
\begin{table*}[!htbp]
\centering  % 表居中
\begin{tabular}{|l|c|c|c|c|c|c|c|c|c|}  % {lccc} 表示各列元素对齐方式，left-l,right-r,center-c
\hline
Method &Anger &Anticipation &Disgust &Fear &Joy &Sadness &Surprise &Trust &Overall\\
\hline
\hline
$\rm{CNN}_{Avg}$   &56.8	&42.5	&58.4	&67.4	&68.5	&64.5	&78.3	&40.3	&-1\\
$\rm{CNN}_{Avg}+\rm{DFT}_{Avg}$   &60.0	&44.7	&63.9	&69.1	&66.2	&59.7	&76.6	&40.9	&-1\\
$\rm{CNN}_{Avg}+\rm{DFT}_{LLC}$   &58.2	&46.9	&60.0	&65.7	&66.2	&66.1	&74.3	&46.1	&-1\\
$\rm{CNN}_{Avg}+\rm{DFT}_{FV}$    &64.4	&37.8	&65.0	&69.8	&57.5	&69.4	&70.6	&47.3	&-1\\
$\rm{CNN}_{Avg}+\rm{DFT}_{VLAD}$  &58.5	&42.2	&64.2	&66.1	&68.2	&62.4	&74.4	&46.4	&-1\\
\hline
\hline
$\rm{CNN}_{LLC}$   &60.2	&46.9	&55.6	&68.3	&62.6	&57.1	&72.4	&48.7	&-1\\

$\rm{CNN}_{LLC}+\rm{DFT}_{Avg}$   &61.8	&43.8	&64.7	&62.0	&67.8	&57.6	&72.7	&53.0	&-1\\
$\rm{CNN}_{LLC}+\rm{DFT}_{LLC}$   &60.9	&48.8	&67.9	&64.1	&64.5	&56.7	&72.3	&52.7	&-1\\
$\rm{CNN}_{LLC}+\rm{DFT}_{FV}$    &63.5	&40.3	&68.7	&67.0	&68.2	&65.2	&80.6	&57.0	&-1\\
$\rm{CNN}_{LLC}+\rm{DFT}_{VLAD}$  &61.2	&45.0	&65.0	&62.8	&65.5	&66.1	&78.6	&55.2	&-1\\
\hline
\hline
$\rm{CNN}_{FV}$    &48.5	&71.6	&68.4	&76.9	&64.3	&67.0	&75.5	&51.8	&-1\\

$\rm{CNN}_{FV}+\rm{DFT}_{Avg}$   &57.1	&55.0	&66.3	&70.9	&78.3	&63.9	&85.6	&52.1	&-1\\
$\rm{CNN}_{FV}+\rm{DFT}_{LLC}$   &62.1	&51.9	&65.5	&77.0	&80.3	&68.2	&85.7	&51.5	&-1\\
$\rm{CNN}_{FV}+\rm{DFT}_{FV}$    &66.8	&62.2	&73.7	&76.1	&67.0	&78.5	&82.1	&55.5	&\textbf{-1}\\
$\rm{CNN}_{FV}+\rm{DFT}_{VLAD}$  &60.6	&50.3	&68.7	&73.7	&77.2	&69.4	&87.8	&51.5	&-1\\
\hline
\hline
$\rm{CNN}_{VLAD}$  &44.1	&49.4	&70.1	&65.1	&54.8	&64.8	&55.3	&56.4	&-1\\

$\rm{CNN}_{VLAD}+\rm{DFT}_{Avg}$   &55.9	&44.4	&68.4	&67.2	&66.0	&60.3	&76.6	&48.8	&-1\\
$\rm{CNN}_{VLAD}+\rm{DFT}_{LLC}$   &58.2	&46.3	&64.2	&65.6	&66.3	&62.1	&72.5	&52.1	&-1\\
$\rm{CNN}_{VLAD}+\rm{DFT}_{FV}$    &65.3	&49.7	&79.2	&74.3	&60.5	&71.5	&72.7	&54.2	&-1\\
$\rm{CNN}_{VLAD}+\rm{DFT}_{VLAD}$  &61.5	&43.8	&67.9	&72.4	&68.3	&62.1	&74.9	&46.1	&-1\\
\hline
\end{tabular}
\caption{Prediction accuracy (\%) of each emotion category using the concatenation of CNN and DFT features using different pooling methods.}
\end{table*}

\begin{table*}[!htbp]
\centering  % 表居中
\begin{tabular}{|c|c|c|c|c|c|c|c|c|c|}  % {lccc} 表示各列元素对齐方式，left-l,right-r,center-c
\hline
Method &Anger &Anticipation &Disgust &Fear &Joy &Sadness &Surprise &Trust &Overall\\
\hline
\hline
Jiang\cite{Jiang2014} &53.0 &7.6 &44.6 &47.3 &48.3 &20.0 &76.9 &28.5 &46.1\\
Pang\cite{pang2015mutlimodal} &50.9 &0.34 &39.9 &54.5 &59.0 &21.7 &82.8 &31.2 &49.9\\
Pang\cite{pang2015deep} &48.5 &0 &53.8 &52.7 &54.2 &32.4 &78.7 &43.8 &51.1 \\
\hline
\hline
$\rm{CNN}_{FV}$    &48.5	&71.6	&68.4	&76.9	&64.3	&67.0	&75.5	&51.8	&\textbf{-1}\\
$\rm{DFT}_{FV}$    &68.2	&37.2	&42.9	&53.0	&33.3	&68.2	&65.6	&33.3	&\textbf{-1}\\
$\rm{CNN}_{FV}+\rm{DFT}_{FV}$    &66.8	&62.2	&73.7	&76.1	&67.0	&78.5	&82.1	&55.5	&\textbf{-1}\\
\hline
\end{tabular}
\caption{Comparison of our results with the three latest works on VideoEmotion-8 dataset.}
%\caption{Prediction accuracy (\%) of each emotion category using CNN, DFT and the concatenation of the two features, and comparison with previous works.}
\end{table*}
\subsubsection{Experimental Results and Discussion}
\textbf{Evaluation of CNN Feature and DFT Features}
The performance of CNN features and DFT features, each with four pooling methods, are shown in Table 1. From Table 1, we find that the performance of CNN features are better than DFT features, with the four pooling methods. For both CNN features and DFT features, FV achieves the best performance, with the accuracy of 65.5\% and 50.2\% respectively, although the dimension of video-level features aggregated by FV is the highest among the four pooling methods. The performances of LLC and VLAD are similar as average pooling.

\textbf{Evaluation of Combining CNN features and DFT features}
%The features generated after pooling method of CNN features and DFT features were normalised to 3$/$5 and 2$/$5 respectively using \ensuremath{\ell_2} normalisation experimentally.
The results of combining CNN features and DFT features, as listed in Table 2, indicate that, by concatenating DFT features, the classification accuracy can be improved. CNN features and DFT features complement each other to achieve satisfactory results.
The concatenation of CNN features with FV and DFT features with FV achieves the best performance 70.2\%.
%When using average pooling for CNN features, the concatenating of DFT features with LLC achieves the best performance. Combining CNN with LLC and , the concatenating of DFT features with LLC achieves the best performance.

%The best performance of For average pooling, LLC, FV and VLAD, the accuracy improved 0.8\%, 4.9\%, 4.7\% and 8.4\% respectively.
%and (3) Compared with LLC, FV and VLAD, the accuracy only improved marginally with average pooling method by combining DFT features.

\textbf{Comparison with State-of-the-arts Results}
In order to demonstrate the effectiveness of our approach, we compare our results with the most recent three works, namely \cite{Jiang2014,pang2015mutlimodal,pang2015deep}.
Comparison results are shown in Table 3, from which we can find:
\begin{enumerate}
  \item[(1)] In comparison to \cite{Jiang2014}, \cite{pang2015mutlimodal} and \cite{pang2015deep}, the accuracy of using CNN features with FV encoding improves 19.4\%, 15.6\% and 14,1\% respectively. In \cite{Jiang2014}, \cite{pang2015mutlimodal} and \cite{pang2015deep}, the authors used low-level visual features, audio features and attribute features, whereas CNN features were applied in our work. Experimental results demonstrate that the performance of CNN features may be superior than hand-crafted features.
  \item[(2)] While using DFT features only can not improve the classification performance, the performance of DFT features with FV encoding is competitive with \cite{Jiang2014,pang2015mutlimodal,pang2015deep}.
  \item[(3)] The highest classification accuracy is 70.2\%, which is obtained by the concatenation of CNN features and DFT features with FV encoding. Combining DFT features achieves 4.7\% better than the performance of using CNN features only. In addition, our best results outperform \cite{Jiang2014} 24.1\%, \cite{pang2015mutlimodal} 20.3\% and \cite{pang2015deep} 19.1\%, which is a significant improvement. To the best our knowledge, our method achieves the best performance at the moment on the VideoEmotion-8 dataset.
%  For four feature pooling methods, combining DFT features with CNN features, performances can be improved, compared to applying CNN features only. Compared to LLC and VLAD, FV achieved the best performance, which outperforms \cite{Jiang2014} 21.7\%, \cite{pang2015mutlimodal} 17.9\% and \cite{pang2015deep} 16.7\%. Overall, incorporating DFT features, the classification performance can be improved 3.7\% compared with applying CNN features with LLC only, 4.4\% compared with applying CNN features with VLAD only, and 2.3\% compared with applying CNN features with FV only.
\end{enumerate}
\subsection{Action Recognition}
\subsubsection{Action Recogntion Dataset}
\textbf{UCF-101 dataset} It consists of 13,320 videos categorised into 101 human action categories with an average of 180 frames per video and a total of 27 hours of video data. Downloaded from YouTube, these videos have fixed frame rate (25 FPS) and resolution ($320\times240$).
Currently, UCF-101 dataset is one of the most challenging datasets for action recognition, due to its large number of categories.
% Our experiments follow the original evaluation scheme using three different training/testing splits.
Following the original evaluation scheme in \cite{jiang2013thumos}, we use three train/test splits.
The average accuracy over the three splits is used to measure the final performance.
%and the mean average precision (MAP) of these splits is reported.
\subsubsection{Implementation Details}
For UCF-101 dataset, two types of features were extracted in our experiments.
\begin{description}
    \item[CNN features] Similar as video emotion recognition, the CNN model pre-trained on ImageNet was adopted for feature extraction. Unlike video emotion recognition, the activation of fc$_6$ were extracted as frame-level features, followed by \ensuremath{\ell_2} normalisation. The dimension of fc$_6$ is 4,096.
    \item[C3D features] For C3D feature extraction, we utilised the public available deep 3-dimension convolutional networks (3D ConvNets) \cite{tran2014learning}, which was pre-trained on I380K and fine-tuned on Sports-1M. ``To extract C3D features, a video is split into 16 frame long clips with a 8-frame overlap between two consecutive clips." The activation of fc$_6$ were extracted as the features, followed by \ensuremath{\ell_2} normalisation. The dimension of fc$_6$ is 4,096.
\end{description}

We trained a vocabulary with 1,024 codewords for LLC, 32 codewords for FV, and 32 codewords for VLAD. The value of $L$ was set to 200 for CNN features and 50 for C3D features. \ensuremath{\ell_2} normalisation was applied for the aggregated CNN features and DFT features. Linear SVM was applied for action recognition. The cost parameter $C$ was set to 1.

\subsubsection{Experimental Results and Discussion}
\begin{table}[!htbp]
\centering  % 表居中
\begin{tabular}{|c|c|}  % {lccc} 表示各列元素对齐方式，left-l,right-r,center-c
\hline
Method &Accuracy(\%) \\
\hline
STIP+BoVW \cite{soomro2012ucf101} &43.9 \\
Deep Net \cite{karpathy2014large} &63.3 \\
CNN + LSTM (Motion) \cite{wu2015modeling} & 81.4 \\
LRCN \cite{donahue2015long} &82.9 \\
%Spatial stream ConvNet \cite{simonyan2014two} &72.6 \\
Temporal stream ConvNet \cite{simonyan2014two} &83.7 \\
Composite LSTM Model \cite{srivastava2015unsupervised} &84.3 \\
C3D (1 nets) + linear SVM \cite{tran2014learning}& 82.3\\
TDD \cite{wang2015action} &90.3 \\
\hline
\hline
$\rm{CNN}_{LLC}$ &52.7\\
$\rm{CNN}_{FV}$ &57.9\\
$\rm{CNN}_{VLAD}$ &51.2\\
$\rm{CNN}_{Avg}$ & 68.8\\

\hline
$\rm{DFT}_{Avg}(CNN)$ &68.8\\
\hline
\hline
$\rm{C3D}_{LLC}$ &73.0\\
$\rm{C3D}_{FV}$ &75.7\\
$\rm{C3D}_{VLAD}$ &61.2\\
$\rm{C3D}_{Avg}$ &82.3\\

\hline
$\rm{DFT}_{Avg}(C3D)$ &82.9\\
\hline
\hline
$\rm{CNN}_{Avg}$+$\rm{C3D}_{Avg}$ &83.7\\
\hline
$\rm{CNN}_{Avg}$+$\rm{DFT}_{Avg}(CNN)$ &\\
+$\rm{C3D}_{Avg}$+$\rm{DFT}_{Avg}(C3D)$ &\textbf{84.1}\\
\hline
\end{tabular}
\caption{Prediction accuracy (\%) of CNN/C3D, DFT features and the concatenation of the two features, and comparison of our results with previous works on UCF-101.}
\end{table}

Action recognition results are shown in table 4.
%The table is divided into four sets. The first set shows state-of-the-arts action recognition results. The rest set shows the results of our approach.
From this table we can find:
\begin{enumerate}
  \item[(1)]The performances of $\rm{DFT}_{Avg}(\rm{C3D})$ and $\rm{DFT}_{Avg}(\rm{CNN})$ are competitive with using CNN features and C3D features.
  \item[(2)] To our surprise, the advanced pooling strategies, e.g. LLC, FV and VLAD, achieve lower accuracy than simple average pooling. This might because that the similarity among frames of UCF-101 videos is much higher than that of VideoEmotion-8 videos.
      %, the average length of videos is much longer and the scenes/backgrouds changes more rapidly.
  %With feature extraction strategy 1, the performances of action recognition were improved by combining CNN and DFT features, compared to using CNN features only. Incorporating DFT features, the recognition performance can be improved  2.6\% compared with applying CNN features with LLC only, 5.6\% compared with applying CNN features with VLAD only, and 2.3\% compared with applying CNN features with FV only.
      %Using DFT features only can not improve the classification performance.
  \item[(3)] The combination of $\rm{CNN}_{Avg}$, $\rm{DFT}_{Avg}(CNN)$, $\rm{C3D}_{Avg}$, $\rm{DFT}_{Avg}(C3D)$ improves 0.4\%, compared to combining $\rm{CNN}_{Avg}$ with $\rm{DFT}_{Avg}(CNN)$. The results demonstrate the effectiveness of DFT features. The improvement of adding DFT features for action recognition is not as significant as that for video emotion recognition. One of the possible reasons might be that combining CNN and C3D features had already achieved satisfactory accuracy. Only small space was left for further improvements.
  \item[(4)] Compared with the state-of-the-art action recognition results, our best result (84.1\%) is competitive. Our result performs 0.2\% worse than composite LSTM model. However, composite LSTM model uses both spatial image features and optical flow features, whereas we only use CNN features and C3D features without using optical flow features. Our method also outperforms LRCN \cite{donahue2015long} 1.2\% and temporal stream ConvNet \cite{simonyan2014two} 0.4\% respectively. Our method achieves lower performance than \cite{wang2015action}. A possible reason might be that we directly apply teh CNN model pre-trained on ImageNet for feature extraction without any fine-tuning. The main purpose of this work is to prove the effectiveness of DFT features rather than challenging the best performance.%Whereas in \cite{wang2015action}, CNN models is trained on

      %trajectory-pooled deep-convolutional descriptors use
  %With feature extraction strategy 2, the improvements of adding DFT feature to CNN features were not significant. One of the possible reasons might be that applying only CNN features had already achieved satisfactory accuracy, i.e over 85\%. Only small space was left for further improvements.
  %\item[(4)]
  %Compare with the state-of-the-arts action recognition results, our best result (86.3\%) is competitive.
\end{enumerate}

\section{Conclusions}
In this paper, we have proposed to analyse features in frequency domain transformed by DFT. In our approach, CNN and DFT features are adopted to jointly model spatial and temporal information for video classification. Capturing temporal information, DFT features have been proved to be efficient and effective for both video emotion classification and action recognition. The combination of CNN and DFT features achieves the state-of-the-art performance on VideoEmotion-8 dataset and competitive results on UCF-101 dataset. %The lower classification accuracy obtained by using static image features, the higher improvements achieved by adding DFT features.% In the future, audio features with its DFT features can be explored for recognizing video emotions.

\section{Acknowledgments}
Haimin Zhang is supported by UTS-CSC international research scholarship.

\bibliographystyle{elsarticle-num}
\bibliography{refs}


%\begin{thebibliography}{00}



%\end{thebibliography}
\end{document}